\documentclass[10pt,twocolumn,letterpaper]{article}

\usepackage{authblk}
\usepackage{cvpr}
\usepackage{times}
\usepackage{epsfig}
\usepackage{graphicx}
\usepackage{amsmath}
\usepackage{amssymb}
\usepackage{subfigure}
\usepackage{enumerate}
\usepackage{amsfonts}

\def\vb#1{\mathbf{#1}}

\def\m#1{\mathsf{#1}}

\usepackage[pagebackref=true,breaklinks=true,letterpaper=true,colorlinks,bookmarks=false]{hyperref}

\cvprfinalcopy 

\def\cvprPaperID{4109} 

\ifcvprfinal\fi
\begin{document}

\title{3D Registration of Curves and Surfaces using Local Differential Information}


\author[1,2]{Carolina Raposo}
\author[1,2]{Jo\~{a}o P. Barreto}
\affil[1]{Institute of Systems and Robotics, University of Coimbra, Portugal\\
\tt\small $\{$carolinaraposo,jpbar$\}$@isr.uc.pt\large}

\affil[2]{Perceive3D, Coimbra, Portugal}

\maketitle

\begin{abstract}
   This article presents for the first time a global method for registering 3D curves with 3D surfaces without requiring an initialization. The algorithm works with {\em 2-tuples point+vector} that consist in pairs of points augmented with the information of their tangents or normals. A closed-form solution for determining the alignment transformation from a pair of matching 2-tuples is proposed. In addition, the set of necessary conditions for two 2-tuples to match is derived. This allows fast search of correspondences that are used in an hypothesise-and-test framework for accomplishing global registration. Comparative experiments demonstrate that the proposed algorithm is the first effective solution for curve vs surface registration, with the method achieving accurate alignment in situations of small overlap and large percentage of outliers in a fraction of a second. The proposed framework is extended to the cases of curve vs curve and surface vs surface registration, with the former being particularly relevant since it is also a largely unsolved problem.
\end{abstract}

\section{Introduction}
\begin{figure}
\centering
\includegraphics[width = 0.9\linewidth]{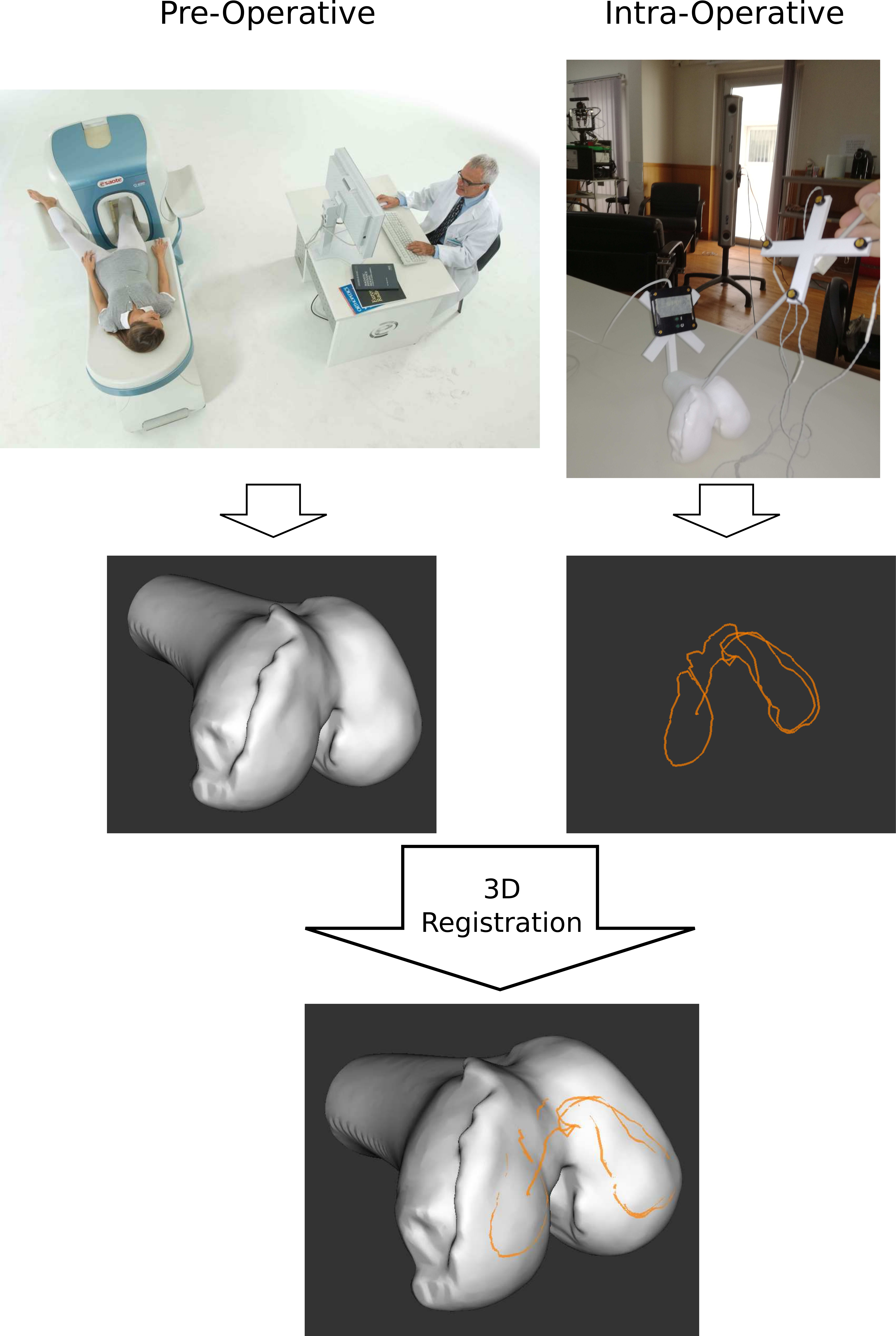}
\caption{3D registration for Computer Aided Orthopaedic Surgery (CAOS).}
\label{fig:schemeintro}
\end{figure}

Finding the rigid transformation that aligns two 3D models is a fundamental problem in computer vision with applications in multiple fields, ranging from robotics~\cite{Pomerleau} to medicine~\cite{almhdie}, and passing by augmented reality~\cite{Wu2013}. This article is motivated by medical applications in general and surgical navigation in orthopaedics in particular~\cite{Mezger}. The workflow of surgical navigation is usually a two step process. First, the surgeon uses a pre-operative 3D image of the targeted anatomy, e.g. a CT or MRI, to plan the procedure. Second, an intra-operative system performs optical tracking of fiducial markers attached to instruments and bones for determining their 3D pose in real-time, that are used to 
guide the surgical execution according to what was established in advance. For this purpose the system must overlay the plan with the actual patient's anatomy in the OR, which passes by using a registration algorithm to align intra-operative 3D data with the pre-operative CT or MRI (Fig.~\ref{fig:schemeintro}).

Thus, 3D registration is a crucial step in surgical navigation and a common approach consists in asking the surgeon to pin-point a number of recognisable anatomical landmarks with a tracked touch-probe. These landmarks are reconstructed in 3D providing explicit point correspondences with the pre-operative image that enable alignment with classical methods~\cite{Horn88}. Unfortunately the approach is not effective in practice because the surgeon often struggles to recognise and access the landmarks, and invariably fails to touch their exact location, which strongly affects the accuracy and robustness of registration. A better alternative is to use as intra-operative 3D data the curves reconstructed by randomly grasping the bone surface with the tracked probe (Fig.~\ref{fig:schemeintro}). However, and to the best of our knowledge, there are no methods in literature to perform the global alignment of a curve and a surface. Stryker~\cite{Stryker} and Exactech~\cite{Exactech} have recently introduced navigation systems that employ randomly reconstructed curves, but they are exclusively used to refine registration using a local ICP variant~\cite{Gruen05,Besl92}, and touching anatomic landmarks is still mandatory to perform initial alignment.  

This article presents for the first time a global method for registering 3D curves with 3D surfaces without requiring a coarse initial alignment. The algorithm works with pairs of points augmented with local differential information that define the so-called {\em 2-tuples point+vector} with the vector being the tangent at the point in case of curves, or the normal at the point in case of surfaces. It is shown that the rigid transformation that aligns curve and surface can be determined in closed-form from a single pair of matching 2-tuples for which the two points correspond. In addition, a 2-tuple point+vector can be described in an invariant manner by a 4-parameter descriptor from which it is possible to derive a set of necessary conditions for a pair of 2-tuples to match. These findings enabled to devise a fast search scheme to establish putative 2-tuple correspondences between curve and surface that are used in an hypothesise-and-test framework to accomplish global registration. 

Comparative tests against plausible alternatives in the literature demonstrate that the proposed algorithm is the first effective solution for global alignment of curves and surfaces. The experiments also show that the approach is well tailored for solving the 3D registration problem in surgical navigation, with the method proving to be very fast and robust, being able to accomplish accurate alignment in situations of small overlap and large percentage of outliers. 

As a final contribution, the framework is extended to the case of registration of curve-vs-curve, which is also a largely unsolved problem, and surface-vs-surface~\cite{Raposo17}. 
\begin{figure}
\centering
\subfigure[Points \emph{vs} Points]{\includegraphics[width = 0.45\linewidth]{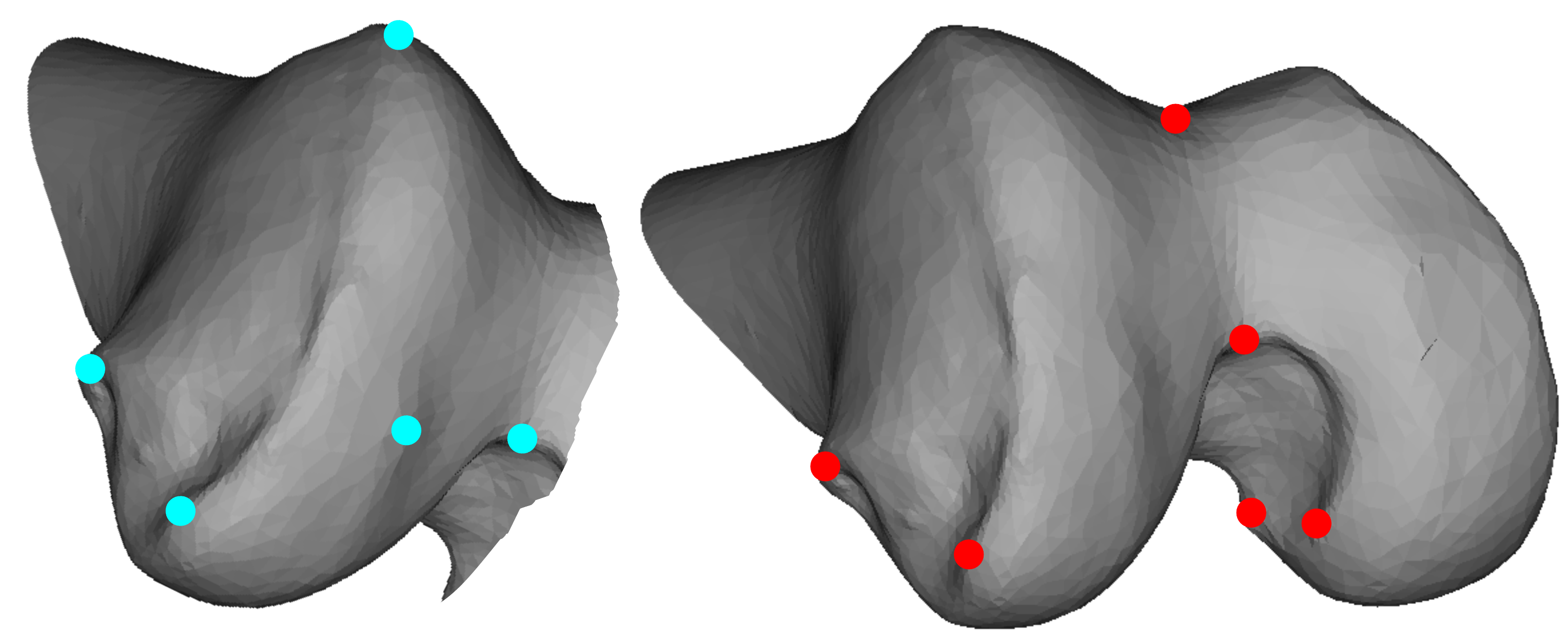}\label{fig:AptsVSpts}} \hfill
\subfigure[Surface \emph{vs} Surface]{\includegraphics[width = 0.45\linewidth]{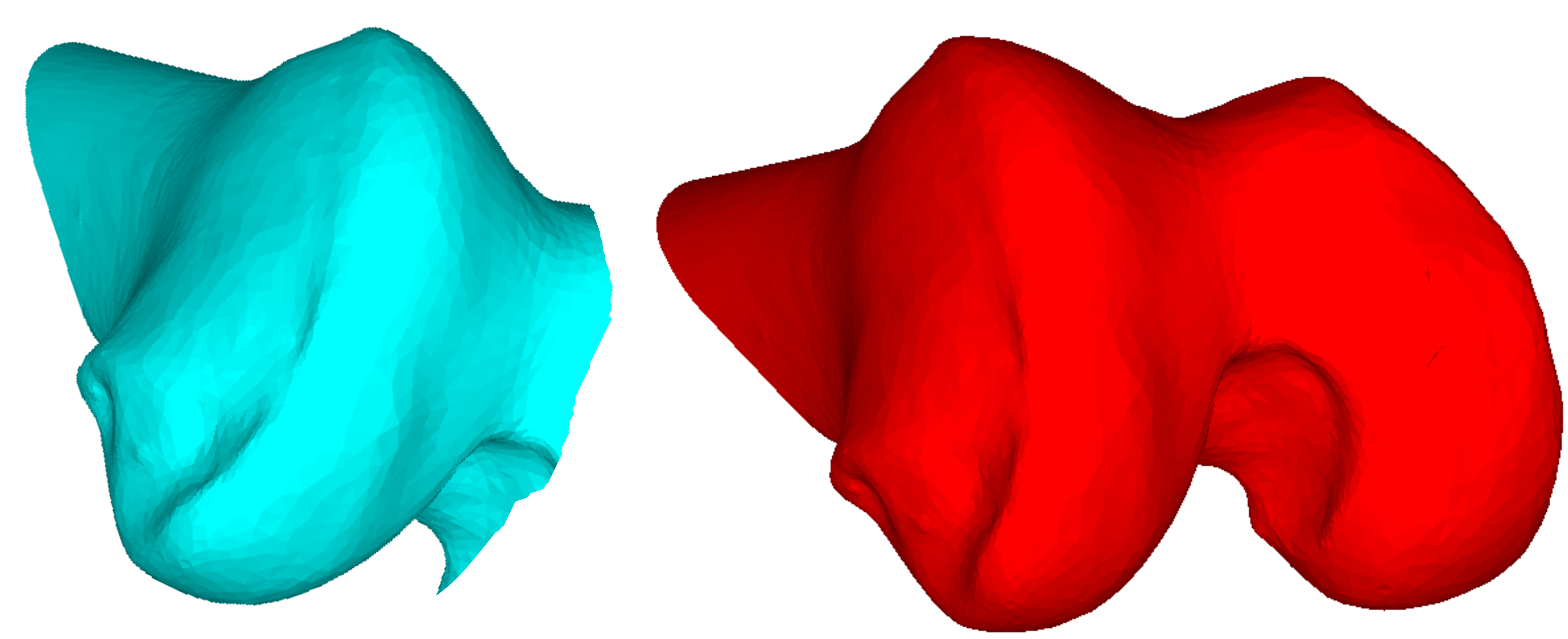}\label{fig:AsurfVSsurf}}

\subfigure[Curves \emph{vs} Surface]{\includegraphics[width = 0.45\linewidth]{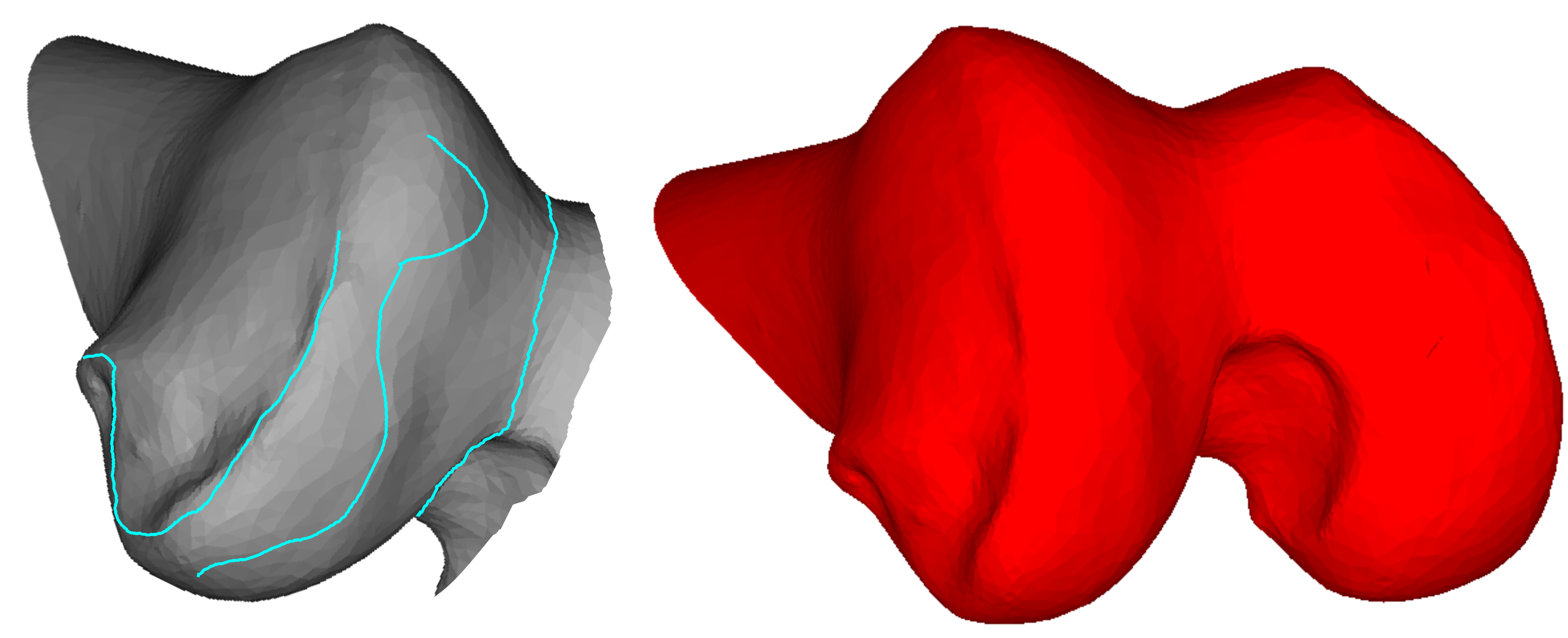}\label{fig:AsurfVSlines}}\hfill
\subfigure[Curves \emph{vs} Curves]{\includegraphics[width = 0.45\linewidth]{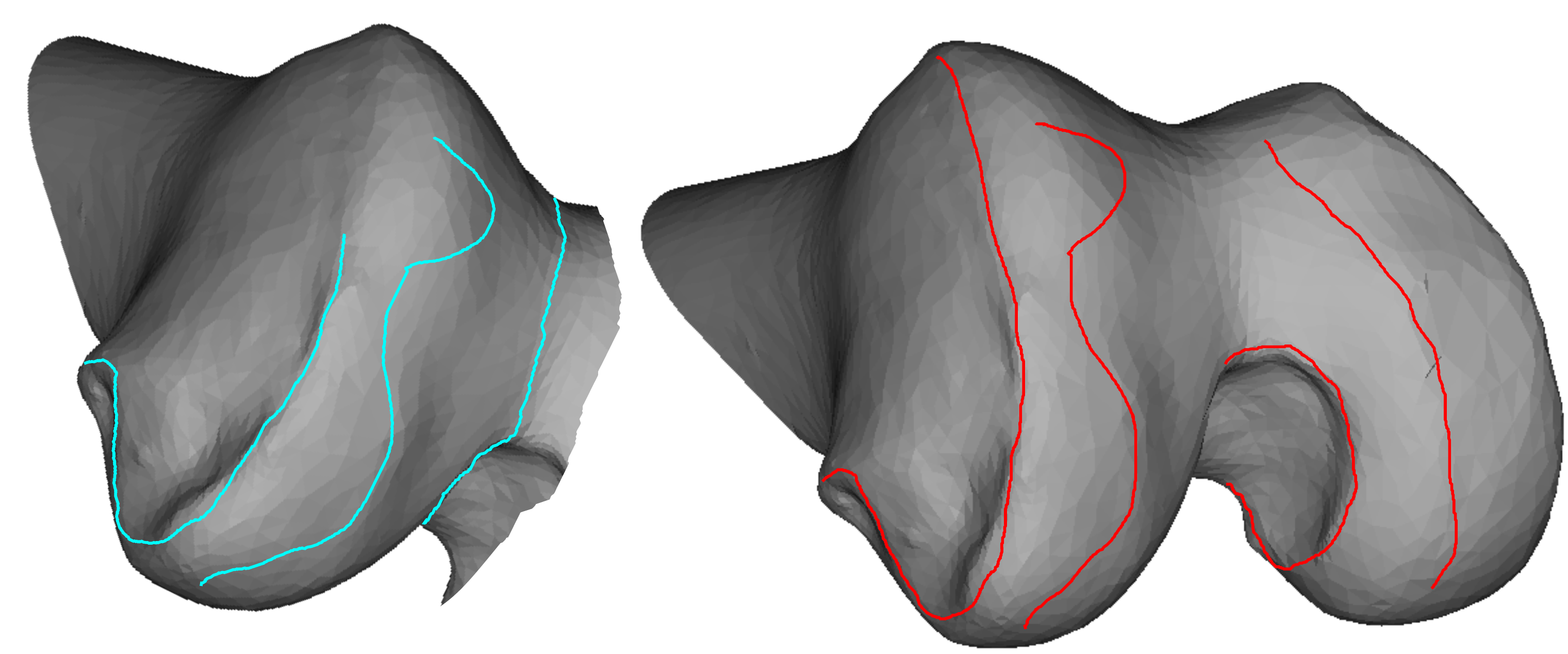}\label{fig:AlinesVSlines}}
\caption{Alternatives of registration using the three existing different types of 3D models: surface, curves and points. This paper provides new solutions for the alignments \subref{fig:AsurfVSsurf}, \subref{fig:AsurfVSlines} and \subref{fig:AlinesVSlines} by making use of differential information.}
\label{fig:3Dcombinations}
\end{figure}

\subsection{Related work}
Despite the vast literature in 3D registration, there is a limited number of authors explicitly addressing curve-vs-surface (Fig. ~\ref{fig:AsurfVSlines}) and curve-vs-curve (Fig. ~\ref{fig:AlinesVSlines}) alignments. Most works concern the alignment of sparse fiducial points (Fig.~\ref{fig:AptsVSpts}) or dense point clouds that are referred to as surfaces (Fig.~\ref{fig:AsurfVSsurf}).   

There are several approaches for the local alignment of two point clouds of which Iterative Closest Point (ICP) is probably the most prominent and broadly disseminated one~\cite{Besl92}. There are numerous variants and modifications of ICP~\cite{almhdie,Pomerleau2013}, many of which work well in situations of refinement of curves-vs-surface or curve-vs-curve registration~\cite{Gruen05}. However, all these methods are local and require proper initialisation to converge. In this article, we aim at global, fast alignment of curves and surfaces, with no prior assumptions about the initial displacement or amount of overlap.

If point correspondences are explicitly known, then the alignment can be accomplished by classical methods without the need of initialization \cite{almhdie}. The recent Go-ICP~\cite{Yang16} and GOGMA~\cite{GOGMA} algorithms use Branch-and-Bound (BB) over the 6-dimensional space of euclidean motions to achieve global registration of Point Clouds without point correspondences. Since the complexity of BB is exponential in the dimension, these methods are slow,  computationally expensive and, from our experiments in curve registration, they often diverge because of small overlap. Several authors propose to handle complexity by searching for rotation and translation separately~\cite{Makadia,Straub_2017_CVPR}. However, the search for the rotation is invariably performed in the space of the surface normals, which precludes the application to curve-vs-surface registration because of the lack of normals in the curve side. 

The 3D surface feature-based algorithms~\cite{Rusu,FGR} involve extracting local features, obtaining matches between features in the two point clouds, and finally estimating the relative pose using RANSAC or other robust estimators. Since curves and surfaces have very different topologies, it is difficult in practice to detect common, coincident saliencies. Moreover many of these methods use feature description for matching which is typically designed for dense point clouds. We run comparative tests with the method of~\cite{FGR} and show that the approach is not amenable for curve-vs-surface registration.

The new family of algorithms 4PCS \cite{amo_fpcs_sig_08,Mohamad15} replaces features by sets of 4 coplanar points whose relations define affine invariants that are preserved under rigid displacements. They work in hypothesize-and-test schemes by selecting a random base of 4 points in the source 3D model and finding all the 4-point sets in the target model that are approximately congruent with the base, i.e. related by a rigid transformation. Despite the search being in linear time, the approach is not suitable for performing curve vs surface alignment because of the very small overlap that dramatically increases search time, as shown by our experiments.

More closely related with our work is the article of \cite{Gourdon94} that also uses local differential information for 3D registration of curves. Two methods are proposed: the first only requires a point correspondence between curves, which considerably decreases the complexity of search, but involves the computation of third order derivatives which is impractical in real, noisy data; the second uses two correspondences, leading to a dramatic increase in the complexity. We present a more clear mathematical formulation of the problem, and provide new insights that lead to an effective search scheme.\\

\textbf{Notation: }
Matrices are represented by symbols in sans serif font, e.g. $\m R$, vectors are represented by bold symbols, e.g. $\vb t,\vb d$, and scalars are indicated by plain letters, e.g. $x, \lambda, N$. Normals and tangents are represented by lower case bold symbols and 3D points are written in upper case bold letters.
\section{Curve vs surface registration using 2-tuples point+vector}\label{sec:registCvsS}
\begin{figure}[!t]
\centering
\includegraphics[width = 0.8\linewidth]{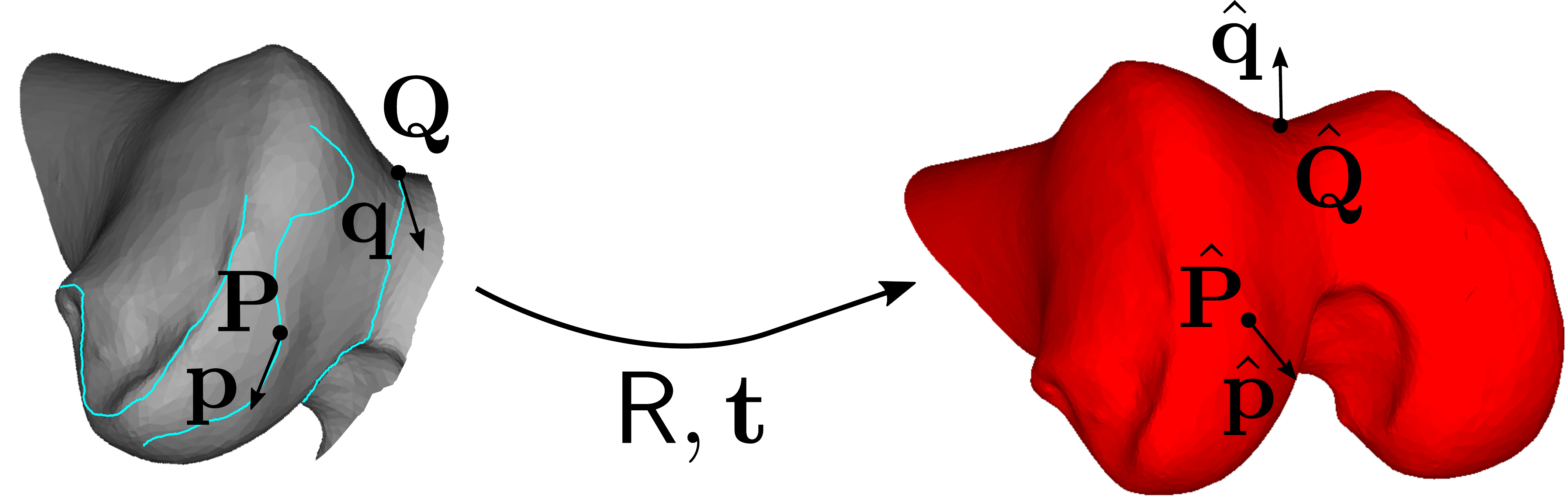}
\caption{Registration of curve C with surface S using a matching 2-tuple point+tangent in C with a 2-tuple point+normal in S.}
\label{fig:pairs_normalstgs}
\end{figure}
This section presents a method for estimating the rigid transformation $\m T$, with rotation and translation components $\m R$ and $\vb t$, respectively, that aligns a curve C with a surface S, as depicted in Fig.~\ref{fig:pairs_normalstgs}. 
For this purpose, we start by showing that it is possible to compute $\m T$ from a pair of corresponding points $\vb P,\vb Q$ and $\hat{\vb P}, \hat{\vb Q}$, together with the information of their tangents $\vb p, \vb q$ on the curve side and their normals $\hat{\vb p}, \hat{\vb q}$ on the surface side. In the remainder of this paper, the pair of points with the corresponding tangents/normals will be referred to as a 2-tuple point+vector and all the tangents and normals in the mathematical derivations are assumed to be unitary.

This section also shows how a 2-tuple point+vector can be described in a compact, translation- and rotation-invariant manner by a 4-parameter descriptor $\Gamma$, and provides the derivation of the necessary conditions for a 2-tuple point+tangent to be a match of a 2-tuple point+normal. These conditions are used in Section~\ref{sec:regist} to effectively establish putative matches that allow a fast 3D registration. 

\subsection{Closed-form solution for curve vs surface registration}\label{sec:estimRt}
\begin{figure}
\centering
\subfigure[$\m R_1 = e^{\left[\vb \omega\right]_\times \alpha}$]{\includegraphics[width = 0.4\linewidth]{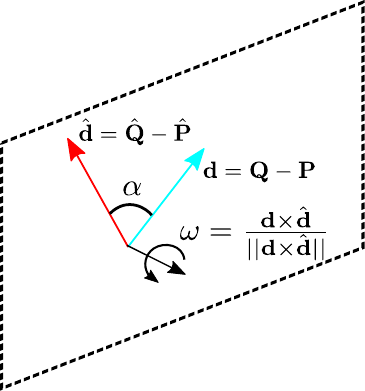}\label{fig:rot1}} \qquad
\subfigure[$\m R_2 = e^{\lambda^{-1}\left[\hat{\vb d}\right]_\times \beta}$]{\raisebox{0mm}{\includegraphics[width = 0.4\linewidth]{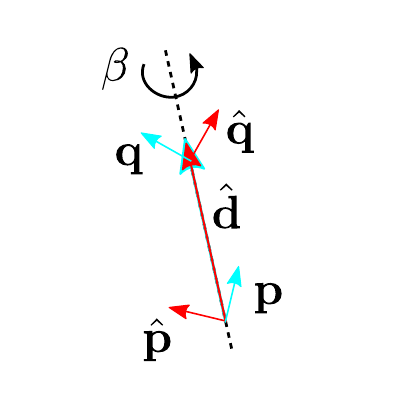}}\label{fig:rot2}}
\caption{The estimation of rotation $\m R$ is divided into the computation of two rotations $\m R_1$ and $\m R_2$.}
\label{fig:rotation}
\end{figure}
Let $\vb P,\vb Q, \vb p,\vb q$ and $\hat{\vb P},\hat{\vb Q}, \hat{\vb p},\hat{\vb q}$ be two corresponding 2-tuples point+vector in curve C and surface S, respectively, and $\m R,\vb t$ the rigid displacement that aligns C with S.
Rotation $\m R$ can be determined independently of translation $\vb t$ as the succession of two rotations: $\m R_1$ that aligns vectors $\vb d = \vb Q-\vb P$ and $\hat{\vb d} = \hat{\vb Q}-\hat{\vb P}$, and $\m R_2$ that places tangents $\vb p,\vb q$ in the planes defined by normals $\hat{\vb p},\hat{\vb q}$, respectively. This can be written as
\begin{equation}
\m R = \m R_2 \m R_1,
\end{equation}
where rotation $\m R_1$ is represented in angle-axis format by
\begin{equation}
\m R_1 = e^{\left[\vb \omega\right]_\times\alpha},
\end{equation}
with $\vb \omega$ being the normal to the plane defined by vectors $\vb d$ and $\hat{\vb d}$, as illustrated in Fig.~\ref{fig:rot1}, and $\alpha$ being given by $\alpha = \cos^{-1}\left(\lambda^{-2}\vb d^\m T\hat{\vb d}\right)$, with $\lambda = ||\vb d|| = ||\hat{\vb d}||$.

Having vectors $\vb d$ and $\hat{\vb d}$ aligned using rotation $\m R_1$, a second rotation $\m R_2$ around $\hat{\vb d}$ by an angle $\beta$ (Fig.~\ref{fig:rot2}) must be performed in order to make vectors $\m R_1\vb p$ and $\m R_1 \vb q$ be orthogonal to $\hat{\vb p}$ and $\hat{\vb q}$, i.e., 
$\m R_2$ must satisfy the following conditions
\begin{equation}
\begin{array}{l c l}
\hat{\vb p}^\m T \m R_2\m R_1 \vb p &=& 0\\
\hat{\vb q}^\m T \m R_2\m R_1 \vb q &=& 0
\end{array}.
\label{eq:systemR2}
\end{equation}
Using Rodrigues' formula, $\m R_2$ can be written as
\begin{equation}
\m R_2 = \m D + \left(\m I-\m D\right)\cos \beta + \lambda^{-1}\left[\hat{\vb d}\right]_\times\sin \beta,
\end{equation}
where $\m I$ is the $3\times 3$ identity matrix and $\m D = \lambda^{-2}\hat{\vb d}\hat{\vb d}^\m T$. Replacing $\m R_2$ in the system of equations~\ref{eq:systemR2} by the previous expression, it comes that $\beta$ can be determined by solving the following matrix equation
\begin{equation}
\m M\begin{bmatrix}
\cos \beta\\
\sin \beta\\
1
\end{bmatrix}=
\begin{bmatrix}
0\\0
\end{bmatrix}, \label{eq:M}
\end{equation}
where $\m M$ is given by
\begin{equation*}
\m M =\\
\begin{bmatrix}
\vb p^\m T\m R_1^\m T\left(\m I -\m D\right)\hat{\vb p} & -\lambda^{-1}\vb p^\m T\m R_1^\m T\left[\hat{\vb d}\right]_\times \hat{\vb p} &  \vb p^\m T\m R_1^\m T\m D \hat{\vb p}\\
\vb q^\m T\m R_1^\m T\left(\m I -\m D\right)\hat{\vb q} & -\lambda^{-1}\vb q^\m T\m R_1^\m T\left[\hat{\vb d}\right]_\times \hat{\vb q} &  \vb q^\m T\m R_1^\m T\m D \hat{\vb q}
\end{bmatrix}.
\end{equation*}
Please note that matrix $\m M$ is not an arbitrary $2\times 3$ matrix. Its structure must be such that the first two values of its right-side null space are consistent sine and cosine values. This idea will be further explored in Section~\ref{sec:conditions}.

Given rotation $\m R$, the translation can be determined in a straightforward manner using one of the point correspondences: $\vb t = \hat{\vb P}-\m R \vb P$.

\subsection{Translation- and rotation-invariant descriptor of 2-tuples point+vector}
\begin{figure}
\centering
\subfigure[Local reference frame]{\includegraphics[width = 0.55\linewidth]{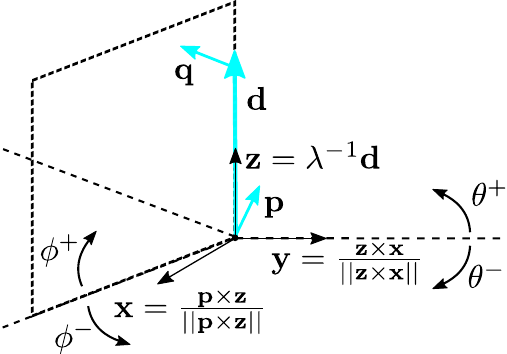}\label{fig:ref}} \hfill
\subfigure[$\Gamma = \left(\lambda, \phi_\vb p, \phi_\vb q, \theta_\vb q\right)$]{{\includegraphics[width = 0.38\linewidth]{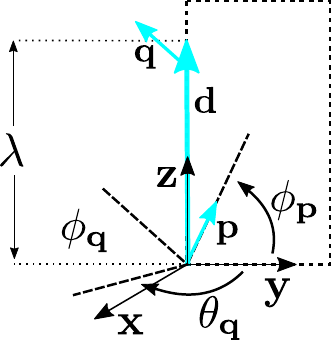}}\label{fig:descript}}
\caption{Representation of \subref{fig:ref} the proposed local reference frame that allows the establishment of a \subref{fig:descript} translation- and rotation-invariant descriptor $\Gamma$. }
\label{fig:sec_descriptor}
\end{figure}

At this point, it is possible to compute $\m R,\vb t$ given matching 2-tuples between a curve and a surface. However, there is still the challenge of, given a 2-tuple in one side, finding potential correspondences on the other side. This section describes a compact description of a generic 2-tuple that will prove to be useful for carrying this search.

Let $\vb P,\vb Q$ be two points and $\vb p,\vb q$ be the corresponding vectors that can either be tangents, in case $\vb P,\vb Q$ belong to a curve, or normals, in case $\vb P,\vb Q$ lie on a surface.

Consider a local reference frame with origin in $\vb P$, with the $\vb z$ axis aligned with $\vb d =\vb Q-\vb P$, and with the $\vb y$ axis oriented such that it is coplanar with vector $\vb p$ and points in the positive direction. This arrangement is depicted in Fig.~\ref{fig:ref}, where $\vb z = \frac{\vb d}{||\vb d||}$, $\vb x = \frac{\vb p \times \vb z}{||\vb p \times \vb z||}$ and $\vb y = \frac{\vb z \times \vb x}{||\vb z \times \vb x||}$.

The local cartesian coordinates can now be replaced by spherical coordinates which are particularly convenient to represent vectors. Choosing these coordinates such that the azimuth of vector $\vb p$ is zero, it comes that the mapping from cartesian $(x,y,z)$ to spherical ($\rho, \theta, \phi$) coordinates is
\begin{equation}
\begin{array}{lcl}
\rho & = & \sqrt{x^2+y^2+z^2}\\
\theta &=& \tan^{-1}-\frac{x}{y}\\
\phi &=& \tan^{-1}\frac{z}{\sqrt{x^2+y^2}}
\end{array},
\end{equation}
where $-\pi < \theta < \pi$ and $-\frac{\pi}{2} < \phi < \frac{\pi}{2}$.

The cartesian coordinates of vectors $\vb d, \vb p$ and $\vb q$ in the local reference frame, expressed in terms of azimuth $\theta$ and elevation $\phi$ are
\begin{equation}
\vb d = \begin{bmatrix}
0 \\
0 \\
\lambda
\end{bmatrix}, \vb p = \begin{bmatrix}
0 \\
\cos \phi_\vb p \\
\sin \phi_\vb p
\end{bmatrix}, \vb q = \begin{bmatrix}
-\sin \theta_\vb q \cos \phi_\vb q \\
\cos \theta_\vb q \cos \phi_\vb q \\
\sin \phi_\vb q
\end{bmatrix},\label{eq:spheric}
\end{equation}
with $\lambda = ||\vb d||$.

Equation~\ref{eq:spheric} emphasizes an important fact that is that an appropriate choice of local frame allows to uniquely describe a 2-tuple point+vector up to translation and rotation using only 4 parameters, which are used to construct vector $\Gamma$ (Fig.~\ref{fig:descript}):
\begin{equation}
\Gamma = \left[\lambda, \phi_\vb p, \phi_\vb q, \theta_\vb q\right]^\m T.
\label{eq: descript}
\end{equation}

Further mathematical manipulation enables to directly move from a 2-tuple $\vb P,\vb Q,\vb p, \vb q$ to its descriptor $\Gamma$ by applying the following vector formulas
\begin{equation}
\begin{array}{lcl}
\lambda &=& ||\vb d|| \\
\phi_{\vb p} & = & \frac{\pi}{2} - \cos^{-1}\left(\frac{\vb {p}^\m T \vb d}{\lambda}\right) \\
\phi_{\vb q} & = & \frac{\pi}{2} - \cos^{-1}\left(\frac{\vb {q}^\m T \vb d}{\lambda}\right)\\
\theta_{\vb q} & = & sign\left(\vb {p}^\m T[\vb d]_\times \vb q\right)\cos^{-1}\left(\frac{\left(\vb q\times \vb d\right)^\m T \left(\vb p \times \vb d\right)}{||\vb q\times \vb d || ||\vb p\times \vb d||}\right)
\end{array},
\end{equation}
where $sign$ represents the signal function.

\subsection{Necessary conditions for a 2-tuple in a curve to match a 2-tuple in a surface}\label{sec:conditions}

Let $\vb P, \vb Q, \vb p, \vb q$ and $\hat{\vb P},\hat{\vb Q}, \hat{\vb p}, \hat{\vb q}$ be 2-tuples in curve C and surface S with descriptors $\Gamma$ and $\hat{\Gamma}$, as defined in Equation~\ref{eq: descript}. If the 2-tuples are not a match, the matrix equation~\ref{eq:M} does not provide a solution with the desired format and rotation $\m R_2$ cannot be estimated.
This section explores this fact to derive the necessary conditions for the pair of 2-tuples $\Gamma$ and $\hat{\Gamma}$ to be a match by enforcing that Equation~\ref{eq:M} has a consistent solution. 

Let $\hat{\Gamma}$ be defined by $\hat{\Gamma} = \left[\hat{\lambda}, \phi_{\hat{\vb p}},\phi_{\hat{\vb q}}, \theta_{\hat{\vb q}}\right]^\m T$. The first condition for $\Gamma$ and $\hat{\Gamma}$ to be a match is that $\lambda = \hat{\lambda}$. Another necessary condition is that there exists a rotation $\m R_2$ that simultaneously makes $\vb p, \vb q$ be orthogonal to $\hat{\vb p}, \hat{\vb q}$. Since we are considering local reference frames for description such that $\vb d$ and $\hat{\vb d}$ are coincident and aligned with a common $\vb z$ axis, the system of equations~\ref{eq:systemR2} becomes
\begin{equation}
\begin{array}{lcl}
\hat{\vb p}^\m T \m R_2 \vb p & = & 0\\
\hat{\vb q}^\m T \m R_2 \vb q & = & 0\\
\end{array}, \text{ with } \m R_2=\begin{bmatrix}
\cos \beta & \sin \beta & 0\\
-\sin \beta & \cos \beta & 0\\
0&0&1
\end{bmatrix}. \label{eq:R2_new}
\end{equation}

Writing $\vb p, \vb q$ and $\hat{\vb p}, \hat{\vb q}$ in terms of the description parameters of $\Gamma$ and $\hat{\Gamma}$, as shown in Equation~\ref{eq:spheric}, and replacing in Equation~\ref{eq:R2_new}, yields
\begin{equation}
\begin{array}{lcl}
\cos \beta &=& -\tan \phi_\vb p \tan \phi_{\hat{\vb p}}\\
\cos(\beta+\theta_{\hat{\vb q}}-\theta_\vb q) &=& -\tan\phi_\vb q\tan\phi_{\hat{\vb q}}
\end{array}.\label{eq:cos}
\end{equation}
Since the cosine varies between $-1$ and $1$, the following must hold to enable the existence of an angle $\beta$:
\begin{equation}
\begin{array}{l c c c l}
-1 &\leq & -\tan\phi_\vb p\tan\phi_{\hat{\vb p}}&\leq& 1\\
-1 &\leq & -\tan\phi_\vb q\tan\phi_{\hat{\vb q}}&\leq& 1
\end{array}.
\end{equation}
Manipulating the previous equations on the elevation angles of descriptors $\Gamma$ and $\hat{\Gamma}$, we obtain a set of inequalities that, together with the distance condition, are necessary conditions for the pair of 2-tuples to be a match:
\begin{equation}
\begin{array}{l c c c l}
& \lambda & = & \hat{\lambda}   &\\
|\phi_\vb p| - \frac{\pi}{2}&\leq & \phi_{\hat{\vb p}}&\leq& \frac{\pi}{2}-|\phi_\vb p| \\
|\phi_\vb q| - \frac{\pi}{2}&\leq & \phi_{\hat{\vb q}}&\leq& \frac{\pi}{2}-|\phi_\vb q|
\end{array}.\label{eq:ineqs}
\end{equation}
\begin{figure}
\centering
\subfigure[No rotation to align]{\includegraphics[width = 0.45\linewidth]{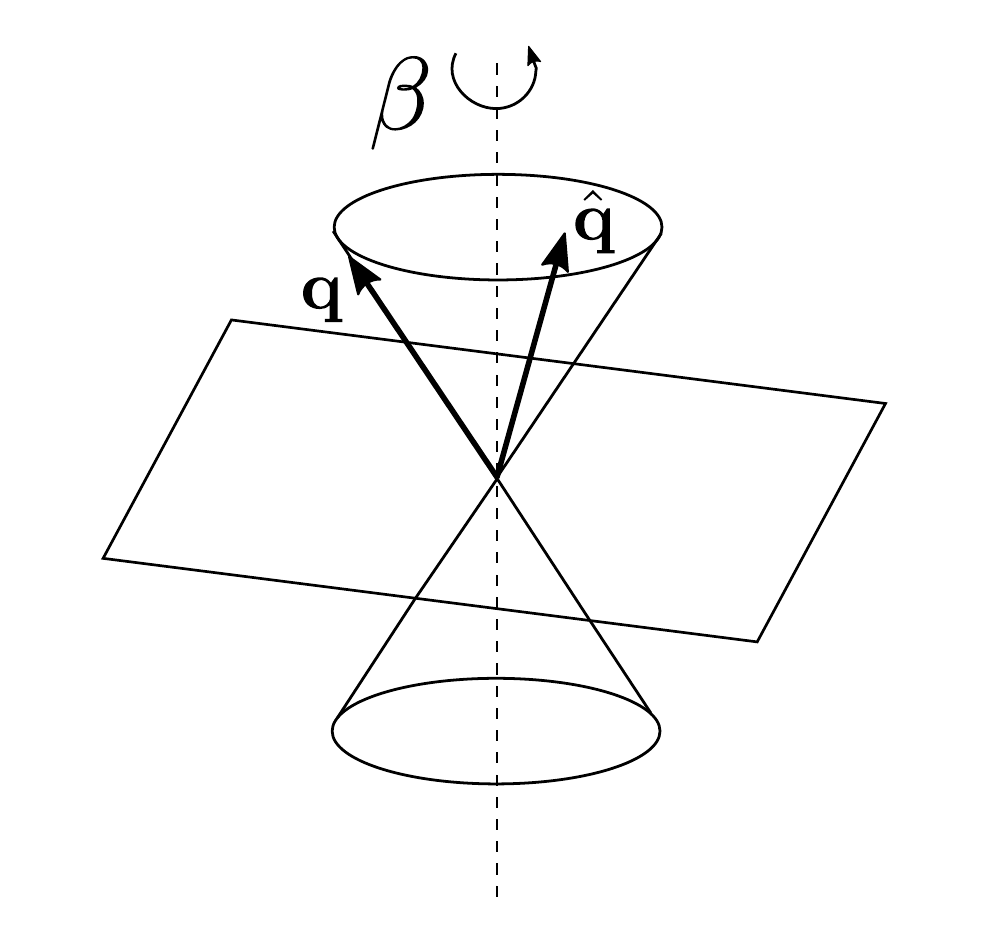}\label{fig:cone1}} \quad
\subfigure[There is a rotation that aligns]{\includegraphics[width = 0.45\linewidth]{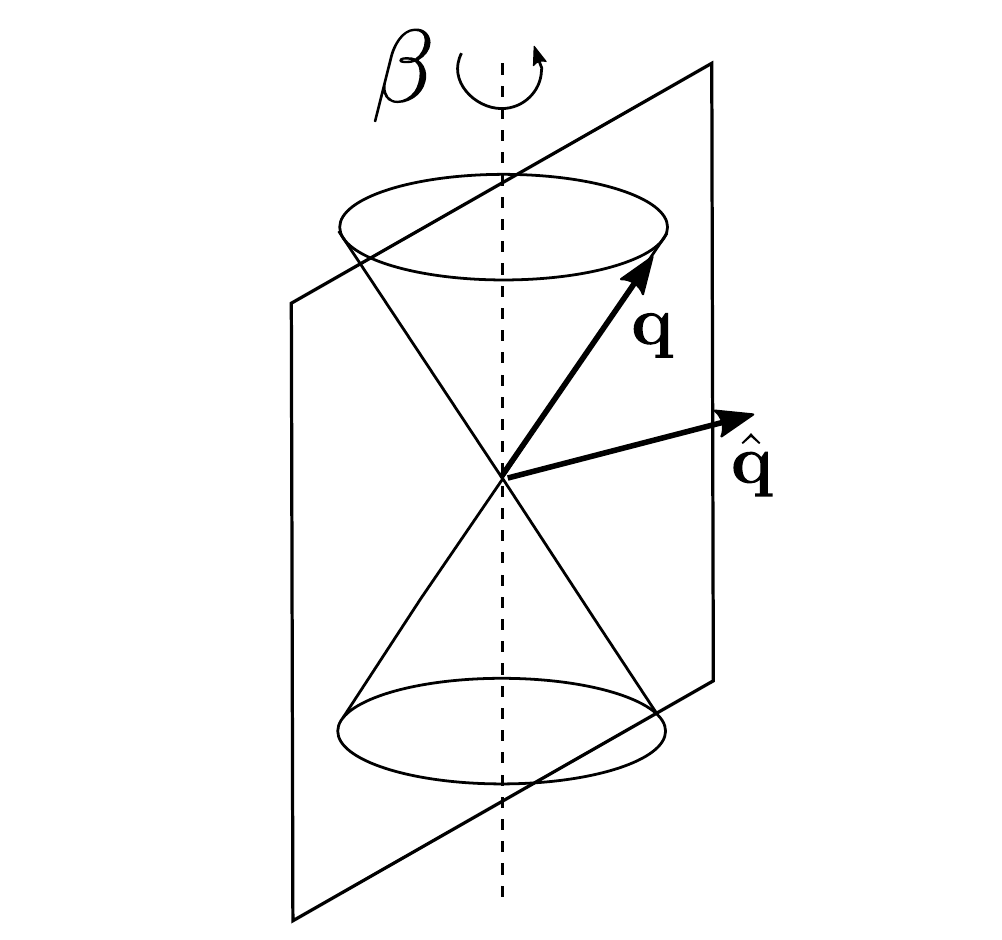}\label{fig:cone2}}
\caption{Condition for tangent $\vb q$ to lie in the plane defined by normal $\hat{\vb q}$.}
\label{fig:cones}
\end{figure}
A careful analysis of the inequalities shows that they are the conditions on the elevation angles for making the cone defined by rotating vector $\vb p$ (or $\vb q$) to intersect the plane defined by $\hat{\vb p}$ (or $\hat{\vb q}$). This is illustrated in Fig.~\ref{fig:cones}, where the two possible situations of non-existence or existence of a valid rotation $\m R_2$ are represented. This figure also clarifies the fact that the orientation of tangents/normals is irrelevant since all the derivations are independent of such orientation.

The previous inequalities must be satisfied in order to exist a rotation $\m R_2$ such that $\vb p$ becomes orthogonal to $\hat{\vb p}$ and $\vb q$ to $\hat{\vb q}$ in separate. A condition on the azimuthal and elevation angles that makes the two pairs of vectors orthogonal in simultaneous can be obtained by manipulating Equation~\ref{eq:cos}:
\begin{equation}
\left(\frac{\tan\phi_{\hat{\vb p}} \tan\phi_\vb p}{\sin\delta_\theta}\right)^2-\left(\delta_\phi^2-2\cos\left(\delta_\theta\right)\delta_\phi+1\right) = 1,\label{eq:reltp}
\end{equation}
with $\delta_\theta=\theta_{\hat{\vb p}}-\theta_\vb q$ and $\delta_\phi = \frac{\tan\phi_{\hat{\vb q}}\tan \phi_\vb q}{\tan\phi_{\hat{\vb p}}\tan \phi_\vb p}$.

If Equation~\ref{eq:reltp} is satisfied, then Equation~\ref{eq:M} has a solution with the desired form.

\section{Method for fast curve vs surface registration}\label{sec:regist}
At this point, and given 2 corresponding 2-tuples, we are able to determine the rigid transformation $\m R,\vb t$. In addition, we proposed a way to describe each 2-tuple by a compact 4-parameter vector, with such description being invariant to translations and rotations, and derived the necessary conditions on these parameters for a 2-tuple $\Gamma$ in curve C to be a potential match of a 2-tuple $\hat{\Gamma}$ in surface S.
The current challenge is in quickly establishing the correspondences such that a fast alignment of the curve and the surface is obtained. This section proposes a solution to this problem.
\begin{figure*}
\centering
\subfigure[Offline processing of the surface.]{\label{fig:scheme1}\includegraphics[width=0.9\linewidth]{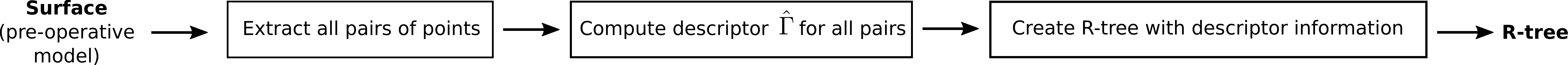}}

\subfigure[Proposed online search scheme that takes advantage of the availability of an offline model.]{\label{fig:scheme2}\includegraphics[width=\linewidth]{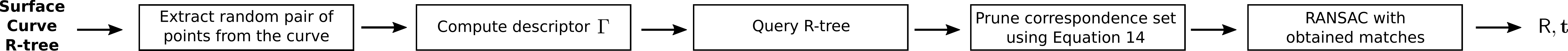}}
\caption[]{Sequences of steps of the proposed \subref{fig:scheme1} offline and \subref{fig:scheme2} online schemes for fast curve vs surface registration.}
\label{fig:schemes}
\end{figure*}

A typical CAOS procedure has an offline stage for obtaining a 3D model of the targeted bone that occurs before the actual surgery is performed. Knowing this, we propose an offline stage for processing the bone model (surface) whose output is used in the online correspondence search scheme, allowing a very fast operation. The sequence of steps of this stage is shown in Fig.~\ref{fig:scheme1}.
The advantage of performing an offline processing of the data is that most of the computational effort of the algorithm is transferred from the online stage to the pre-processing, where computational time is irrelevant. 

We propose to build a data tree structure that contains the relevant information for all pairs of points in order to facilitate and accelerate the online search stage. Firstly, all 2-combinations of points are extracted and their 4-parameter vectors $\hat{\Gamma}$ are computed. Then, a 3-dimensional R-tree is created using all points $(\lambda, \phi_{\hat{\vb p}},\phi_{\hat{\vb q}})$ and $(\lambda, -\phi_{\hat{\vb p}},-\phi_{\hat{\vb q}})$, to account for the switched point-wise correspondences. Each object of the tree also includes the value for $\theta_{\hat{\vb q}}$ and two indices $i,j$ that identify the pair of points in the point cloud.

Our proposed online search scheme (Fig.~\ref{fig:scheme2}) starts by extracting a random pair of points from the curve, and its tangents, and computing its descriptor $\Gamma$. This pair is then used for querying the R-tree for selecting all pairs in the surface that simultaneously have a distance $\lambda\pm\epsilon$, where $\epsilon$ is a parameter to account for noise in the data, and satisfy the conditions in Equation~\ref{eq:ineqs}. The obtained set of pairs is afterwards pruned by choosing only the ones that satisfy Equation \ref{eq:reltp}. The obtained correspondences of pairs of points are then processed in a RANSAC scheme in order to find the rigid transformation that yields the highest number of inliers. If all the correspondences have been processed and the stopping criteria was not met, the algorithm repeats this process for a new random pair of points extracted from the curve.


\section{Extensions to curve vs curve and surface vs surface}\label{sec:registcurve}

This section shows how to solve the global 3D registration
problem for two 3D models of the same type: two curves or two surfaces. We will provide an explanation for the curve vs curve alignment, with the derivations being identical for the case of surface vs surface registration.

Consider two curves C and \^{C} and two corresponding 2-tuples point+tangent with descriptors $\Gamma$ and $\hat{\Gamma}$. The methods and derivations of Section~\ref{sec:registCvsS} hold with two differences. First, the constraints for determining the angle $\beta$ of rotation $\m R_2$ (Equation~\ref{eq:systemR2}) become
\begin{equation}
\begin{array}{ccc}
\hat{\vb p} &=& \m R_2\m R_1 \vb p\\
\hat{\vb q} &=& \m R_2\m R_1 \vb q
\end{array},
\end{equation}
meaning that $\m R, \vb t$ can be computed in closed form using two points and just one tangent. This is valid because in this case $\m R_2$ is the rotation that aligns the corresponding tangent vectors.

The second difference is that the necessary conditions of Equations~\ref{eq:ineqs} and~\ref{eq:reltp} 
become
\begin{equation}
\begin{array}{l c l}
\lambda &=& \hat{\lambda}\\
\phi_\vb p&=&\pm\phi_{\hat{\vb p}}\\
\phi_\vb q&=&\pm\phi_{\hat{\vb q}}\\
\theta_\vb q&=&\theta_{\hat{\vb q}}
\end{array},
\label{eq:systemsol}
\end{equation}
where the $\pm$ sign accounts for the fact that the tangents are in general non-oriented.
Instead of being inequalities, as in the curve vs surface alignment, in this case the conditions for matching are equalities, enabling search mechanisms other than R-trees. This is validated in the experimental section, where the search is carried by extracting the pairs of points that satisfy the conditions in Equation~\ref{eq:systemsol} using the pair extraction scheme proposed in~\cite{Mellado14} that runs in $\mathcal{O}(N)$ time, with $N$ being the number of points in the target curve. Note that this scheme does not contain an offline processing stage as the search algorithm proposed in Section~\ref{sec:registcurve}.

As stated before, the surface vs surface registration is similar to curve vs curve, having the difference that tangents are replaced by normals. We will not discuss this problem further because the resulting method is equivalent to the one that has been recently presented and tested in~\cite{Raposo17}.

\section{Experiments}\label{sec:experiments}
This section reports tests performed on synthetic and real data in order to assess the accuracy and speed of the proposed registration methods. The first experiments use synthetic data for which the ground truth rigid transformations are known, and compare our curve vs surface and curve vs curve registration methods with two state-of-the-art approaches for which there is public implementation available: Super4PCS~\cite{Mellado14} and Fast Global Registration (FGR)~\cite{FGR}. The last experiment attempts to mimic a common CAOS procedure, where 3D data on the surface of a bone is reconstructed and registered with a pre-operative virtual dense model of that bone.

The normals were computed using the PlanePCA algorithm~\cite{planePCA} with a neighbourhood of 30 points and the tangents were estimated using a standard algorithm for computing Frenet frames. 
Both registration algorithms were implemented in C++ and all tests were performed on a Intel Core i5-6200U CPU @ 2.30GHz with 8GB of RAM.
\subsection{Curve vs surface registration using synthetic data}\label{sec:cVSs}
\begin{figure}
\centering
\includegraphics[width=\linewidth]{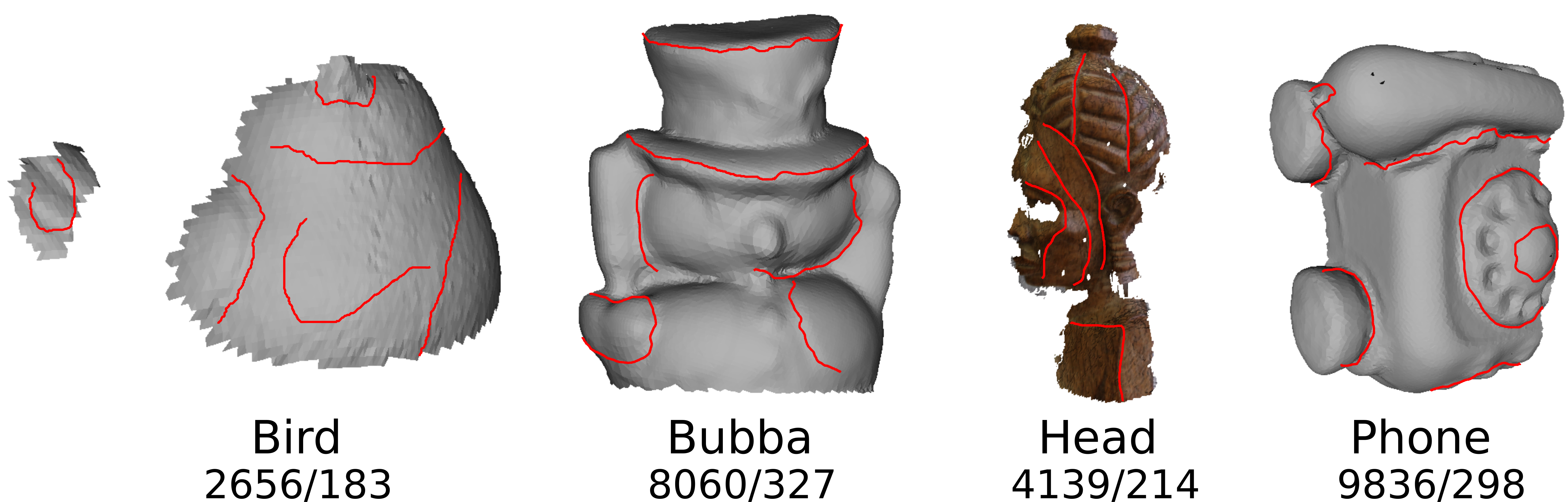}
\caption[]{Models and curves used in the synthetic experiments. Below each model, the two values correspond to the number of points of the model and of the set of curves.}
\label{fig:models}
\end{figure}

In this experiment we used the 4 synthetic models shown in Fig.~\ref{fig:models} for evaluating the performance of the proposed curve vs surface algorithm. The sets of 6 segments shown in red in the figure represent the curves that were manually extracted from each model in order to create the curves for performing the registration.

The extracted sets of curves were used for generating smaller sets of curves with the intent of assessing the performance of the algorithms for different amounts of data.
This was done by randomly choosing 2 and 4 out of the 6 segments and then selecting a random set of contiguous points in each segment such that the total number of points is about $25\%$ and $50\%$ of the total number of points of the original sets, respectively.
This scheme was used for creating 25 different sets of each of the two sizes, to which random rigid transformations are applied. We also consider the full curve (containing $100\%$ of the points) in 25 different initial poses generated randomly. This procedure yields 75 different input curves for each model, to which random noise drawn from the standard normal distribution with mean 0 and standard deviation $\sigma$ is added. We test both with noise-free data ($\sigma = 0$) and by adding noise with $\sigma = 1$ which represents $1.5\%$ of the diameter of the models, whose dimensions were previously adjusted by setting their diameters to $75$. The noise was added to each point independently. This level of noise ($\sigma = 1$) causes the direction of tangents to vary with respect to $\sigma = 0$ by an average of $\approx 9^\circ$. Also, we defined as stopping criteria for the search algorithms a maximum execution time of 5 sec or a percentage of inliers of 95$\%$.

\begin{figure}
\centering
\includegraphics[width=0.95\linewidth]{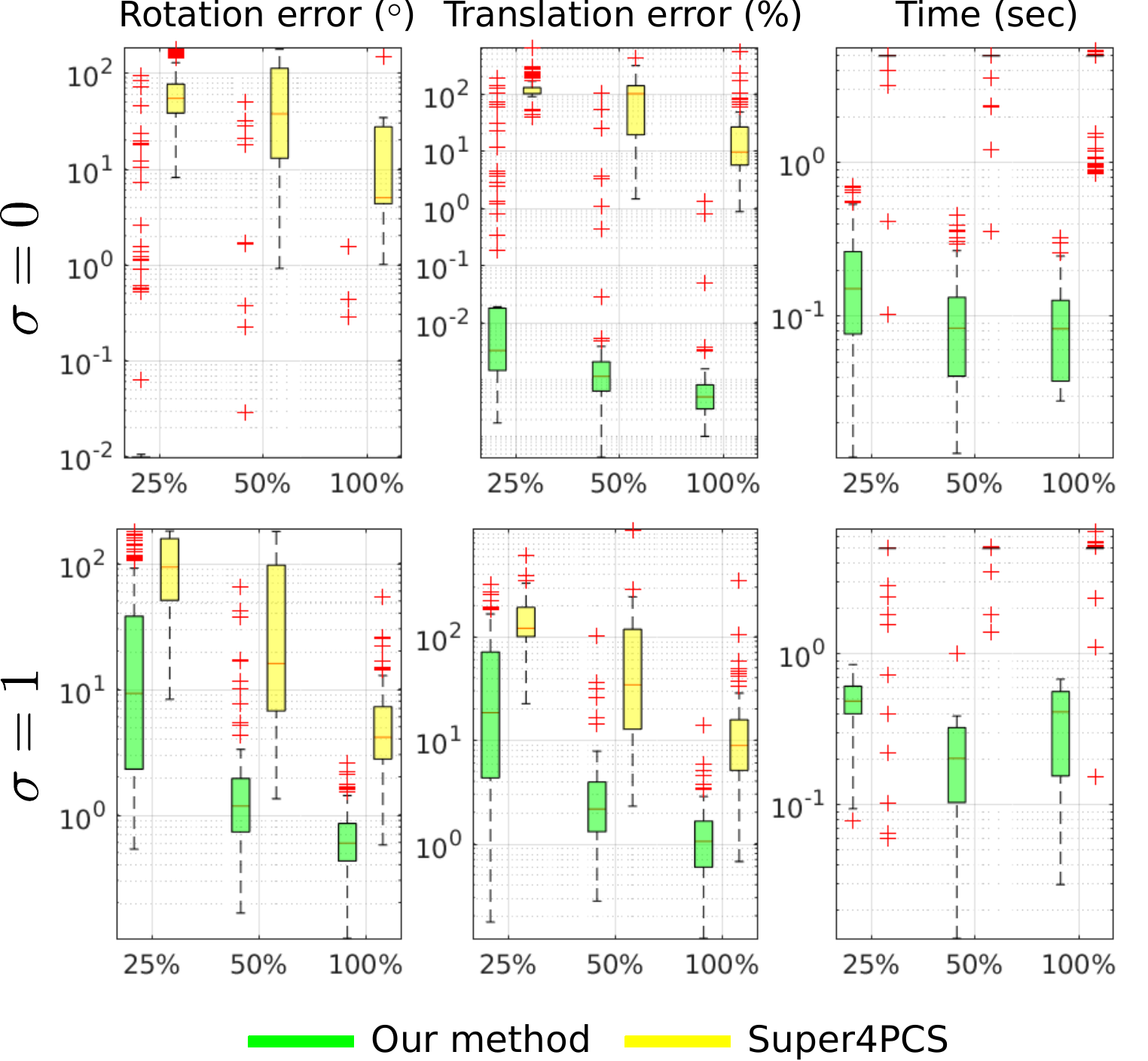}
\caption[]{Results obtained in the curve vs surface registration with our method and Super4PCS for different levels of noise and amount of input data.}
\label{fig:resCvsS}
\end{figure}

Fig.~\ref{fig:resCvsS} shows the results obtained with our proposed curve vs surface method and Super4PCS. We also tested with the FGR method but it failed in all cases because it is a feature-based approach and we are working with two different types of 3D data. Results are given as rotation and translation errors, computed as in~\cite{Raposo17}, and computational times. The results are merged for all models such that each boxplot corresponds to 100 registrations. Regarding our method, the reported times only include the online search stage and information on the offline processing of the models can be found in Table~\ref{tab:times}.

The superiority of our approach w.r.t. Super4PCS both in terms of accuracy and speed is evident as it was able to provide proper alignments for all the different conditions of size of input data and noise. On the other hand, Super4PCS performed poorly for the sets of curves with $25\%$ and $50\%$ of the points of the original data, being only able to provide acceptable solutions for the complete sets. As expected, the accuracy of our method also increases with the amount of input data as a larger coverage of the surface is given. However, even for the smallest curve sizes, it provides good alignments, indicating that the method is able to work with very local information. Concerning computational times, our method is very fast, being able to perform the search in less than 1 sec in all cases. Even if we consider the time corresponding to the offline processing of the surface, which in the worst case is 1.9 sec, the total execution time of our method is still far below Super4PCS's.

\subsection{Curve vs curve registration using synthetic data}\label{sec:cVSc}
\begin{table}[]
\centering
\caption{Computational times of the offline stage for each model.}
\label{tab:times}
\begin{tabular}{|c||c|c|c|c|}
\hline
Model & Bird & Bubba & Head & Phone \\ \hline
Time (sec) & 0.3 & 1.2 & 0.5 & 1.9 \\ \hline
\end{tabular}
\end{table}
\begin{figure}
\centering
\includegraphics[width=0.95\linewidth]{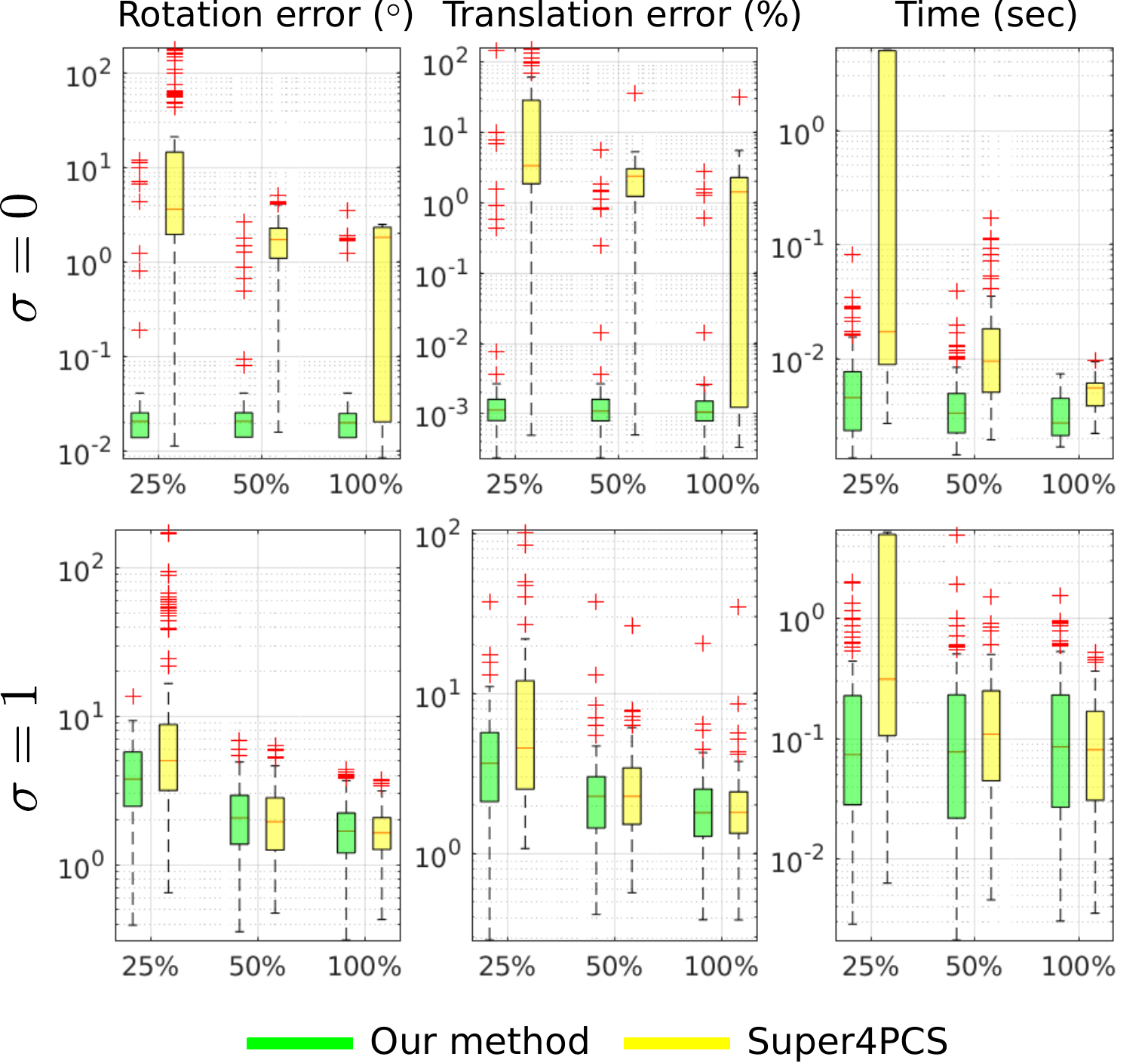}
\caption[]{Results obtained in the curve vs curve registration presented as in Fig.~\ref{fig:resCvsS}.}
\label{fig:resCvsC}
\end{figure}
This experiment is performed similarly to the previous one, having the difference that we perform curve vs curve alignment by replacing the surfaces with the curves represented in Fig.~\ref{fig:models}. We tested with FGR but do not show the obtained results since it was only able to provide acceptable solutions in the noise-free case. When noise was added, the number of corresponding features became very low and the method performed poorly for all curve sizes. 
\begin{figure*}
\centering
\subfigure[Different regions of the knee]{\includegraphics[width = 0.25\linewidth]{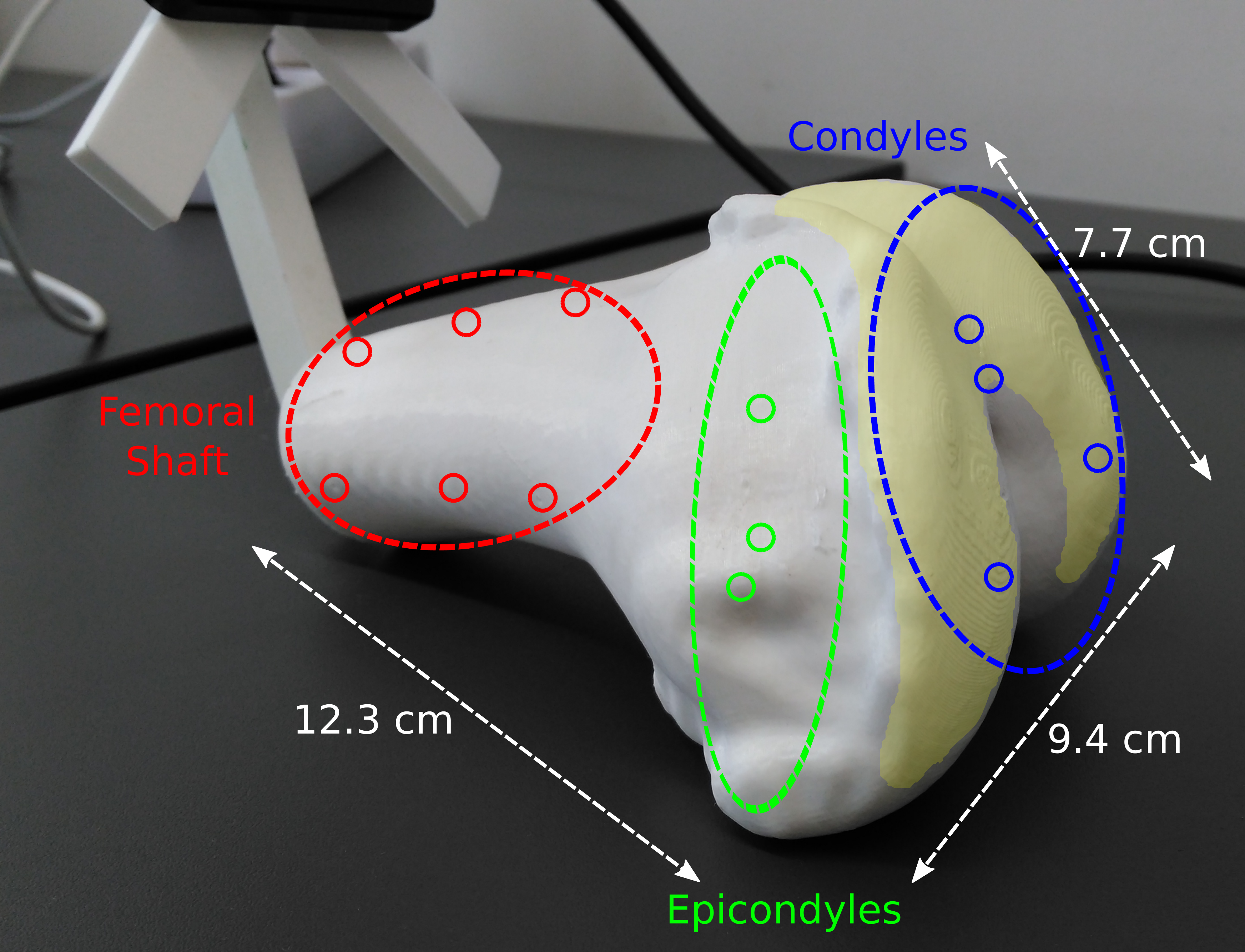}\label{fig:kneeA}} \hfill
\subfigure[Example of a registered curve]{\includegraphics[width = 0.235\linewidth]{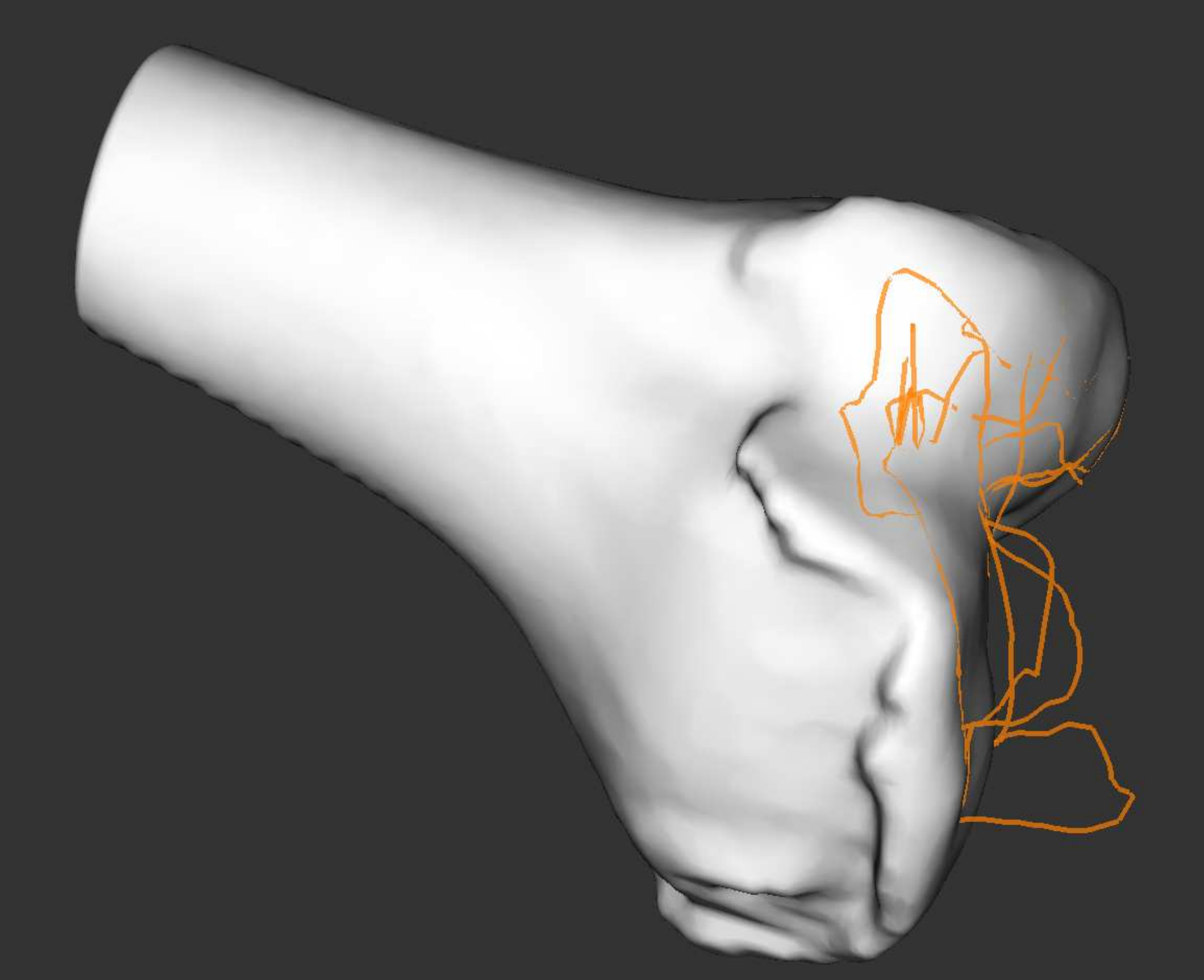}\label{fig:kneeB}}\hfill
\subfigure[Acquisition of control points]{\includegraphics[width = 0.21\linewidth]{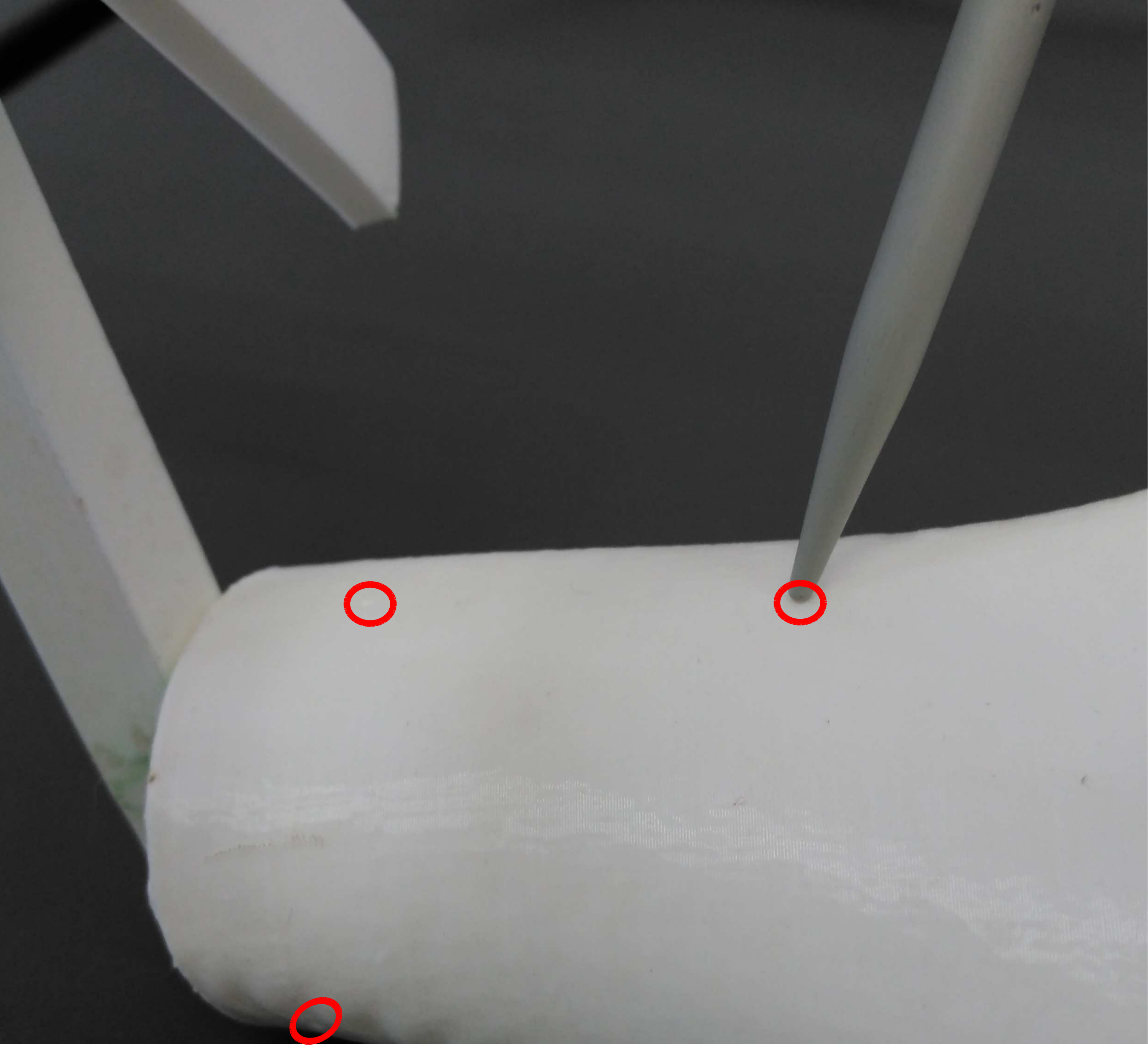}\label{fig:kneeC}}\hfill
\subfigure[Registration results]{\raisebox{-2mm}{\includegraphics[width = 0.29\linewidth]{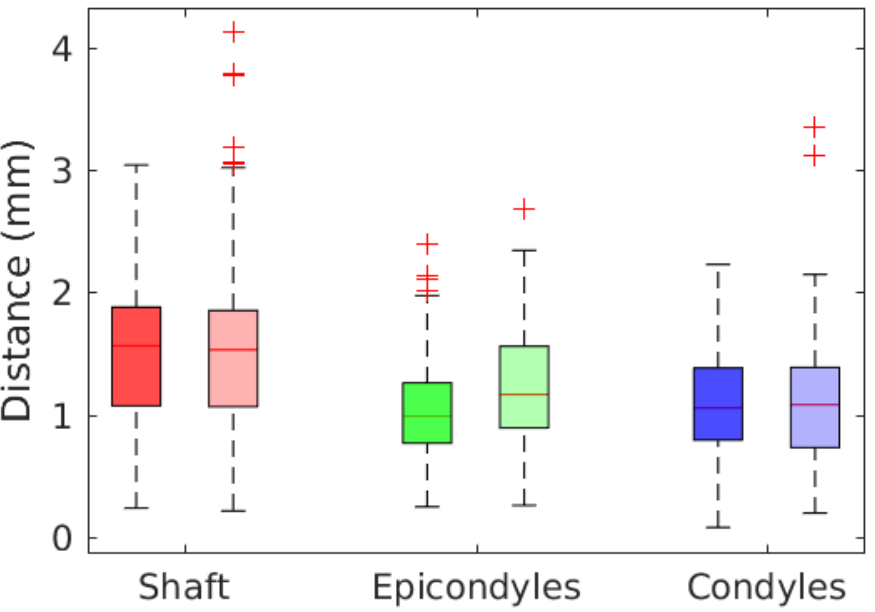}}\label{fig:kneeD}}
\caption{Experiment to mimic a CAOS procedure. \subref{fig:kneeA} Different regions of the knee are identified with different colors and \subref{fig:kneeD} the distances obtained for each region are identified with the same colors. The results obtained with the data containing outliers are shown in transparent boxplots. \subref{fig:kneeB} Example of a curve acquired with outliers and registration result. \subref{fig:kneeC} The 3D coordinates of the control points are obtained by carefully reconstructing points with an instrumented touch probe.}
\label{fig:knee}
\end{figure*}

Results in Fig.~\ref{fig:resCvsC} show the superiority of our approach w.r.t. Super4PCS in the noise-free case. When noise is added, our method still outperforms Super4PCS when the amount of input data is very small, performing similarly when $100\%$ of the data is provided. This can be explained by the fact that tangents are more affected by noise than points, leading to a degradation in the performance of our approach. However, it still manages to accomplish a proper alignment of the curves in all situations in under 1 sec.

\subsection{Experiments in CAOS using a dry knee model}

The last experiment mimics a common procedure in CAOS where 3D data is reconstructed by touching the surface of the bone with an instrumented touch probe and subsequently registered with a pre-operative 3D virtual model of that bone. In order to simulate this, we used a dry knee model to work as the bone, as shown in Fig.~\ref{fig:kneeA}, and acquired 3D curves on the surface of the bone using a state-of-the-art optical tracking system (the Optotrak Certus). We acquired 30 curves in the condyliar region, highlighted in yellow in Fig.~\ref{fig:kneeA}, where 15 of them contained $20\%-40\%$ of outliers. An example of a curve with outliers is given in Fig.~\ref{fig:kneeB}. Each curve was used for performing curve vs surface registration using our proposed approach and the obtained rigid transformations are used to represent the 23 control points illustrated in Fig.~\ref{fig:kneeA} in the virtual model reference frame. These control points had been previously reconstructed by carefully placing the touch probe in the small holes of the model (Fig.~\ref{fig:kneeC}) and their ground truth 3D coordinates in the virtual model reference frame are known.
Fig.~\ref{fig:kneeD} shows the distributions of distances between the transformed and the ground truth points for three different regions of the knee, both for the curves without forced outliers (solid color) and with outliers (transparent).

As expected, the obtained distances are smaller in the region where the data was acquired (condyles). However, in the other regions the errors do not increase substantially, not even in the femoral shaft that is more than 10 cm away from the area of acquisition. This is an important result since it confirms the previous observation that our method is able to properly register large surfaces with very local information, being advantageous in CAOS procedures where the area of the bone that is exposed is often restricted. Another relevant observation is that our method is able to deal with large amounts of outliers, shown by the fact that there was not a significant degradation of the registration accuracy when using data with outliers. This demonstrates that besides being accurate and fast, the proposed method is robust and resilient to outliers, which highly improves its usability.

\section{Conclusions}
We present the first method for fast global registration of curves and surfaces that does not require an initial coarse alignment. The method makes use of pairs of points, augmented with their local differential information, not only to solve the rigid transformation estimation problem but also to establish correspondences of pairs of points in a very fast manner. Experiments demonstrate that the proposed method significantly advances the state-of-the-art by providing a fast and robust 3D registration algorithm that dramatically outperforms two plausible alternatives for global registration. 

As future work, we intend to extend the rigid transformation estimation algorithm to a more general method for determining not only the rotation and translation but also the scale. This has applications in CAOS has it would, for instance, allow problems of difference in size between the virtual models and the respective bones to be overcome.
\section*{Acknowledgements}
The authors thank the Portuguese Science
Foundation and COMPETE2020 program for generous funding
through project VisArthro (ref.: PTDC/EEIAUT/3024/2014). This paper was also funded by the European Union's Horizon 2020 research and innovation programme under grant agreement No 766850.
{\small
\bibliographystyle{ieee}
\bibliography{sample}
}

\end{document}